\documentclass[preprint]{JASA}
\AtBeginDocument{\let\linenumbers\relax\nolinenumbers}
\usepackage{natbib}
\usepackage{graphicx} 
\usepackage{tipa}
\usepackage[margin=1in]{geometry}

\begin{document}

\title{Articulatory strategy as a source of variation in acoustic vowel dynamics}
\author{Patrycja Strycharczuk}
\email{patrycja.strycharczuk@manchester.ac.uk}
\affiliation{Linguistics and English Language, University of Manchester, United Kingdom}
    
\author{Justin J. H. Lo}
\author{Sam Kirkham}
\affiliation{Linguistics and English Language, Lancaster University, United Kingdom}

\date{\today}

\begin{abstract}
        Acoustic vowel dynamics have some speaker-identifying characteristics, which have been ascribed to individual properties of articulatory strategies: formant transitions have a particular shape because speakers move their articulators, using specific and  practised movements. However, there is little existing evidence that different articulatory strategies systematically affect formant dynamics. The present study corroborates the link between the two. Ultrasound tongue imaging data from 36 speakers of Northern-Anglo English are used to identify distinct articulatory strategies for the production of palatal vowel /i/. Tongue shape in /i/ is found to be a significant predictor of formant dynamics in diphthongs with a palatal offglide. The observed relationships can be explained by the characteristics of articulatory movement conditioned by vocal tract shape. Greater articulatory displacement of tongue root and/or dorsum produces greater distortion from the mean tongue shape in palatal vowels, and it also requires higher articulatory velocities, resulting in relatively earlier and steeper formant transitions. The results contribute to the conceptual understanding of individuality in speech, by illuminating the regularising and individual aspects of articulatory compensation.
        \end{abstract}
        
    \maketitle
    
\section{Introduction}
\subsection{Speaker-specificity of vowel dynamics}
\label{specificity}
Vowel dynamics are highly individual. Multiple studies show that measures of inherent change in formant values contain some speaker-identifying information, and the performance of speaker-recognition models tends to improve when dynamic information is included, compared to models based on static formant measurements \citep{mcdougall2006, morrison2009, rhodes2012, rose2006, hughes2016, heeren2020}.  Similar improvements are observed when acoustic information is enriched with articulatory information \citep{kirchhoff2002}. Furthermore, it has been observed that the contribution of dynamic information to speaker-specificity is relatively greater for more diphthongal vowels, and more limited when there is less inherent change within the vowel \citep{hughes2016, heeren2020}. 

A prominent line of explanation for the speaker-specific characteristics of formant dynamics focuses on individual differences in articulatory movement used to achieve specific acoustic targets \citep{nolan2005, mcdougall2004, mcdougall2006, mcdougall2007}. According to \citeauthor{nolan1983}'s \citeyearpar{nolan1983} model of sources of between-speaker variation, given the same set of phonetic representations, speakers have the flexibility to reach their primary auditory goals through different implementational rules, which are themselves informed and constrained by the configurational and dynamic properties of the vocal tract and articulators. The articulatory strategy for vowel production varies from speaker to speaker \citep{johnson1993}, reflecting both anatomical differences and practised articulatory routines. It has been hypothesised that these types of individual characteristics are more directly observable in movement between articulatory targets. Possible reasons for the increased variability of transitions include the perceptually privileged status of targets \citep{mcdougall2007, morrison2009}, and quantal properties of speech \citep{stevens1989}, whereby phonological systems preferably select phonetic variants that are characterised by acoustic stability in the presence of articulatory variance. The same preference does not extend to movements between targets, hence the prediction that transitions will allow for greater acoustic variability \citep{nolan2005}.

The hypothesis that articulatory strategy contributes to individual variation in acoustic vowel dynamics has inspired a large body of forensic phonetic research, but it has not been directly tested. Some indirect support for the articulatory basis of individual variation in vowel dynamics come from phonetic studies of identical twins. 
\citet{weirich2012} reports different levels of similarity between monozygotic and heterozygotic twins, and that these interact with the type of measure: static vs. dynamic. Greater similarity is found for monozygotic twins, but only for dynamic measures (consonant-vowel transitions), and not for static measures. These findings can be attributed to a shared articulatory strategy in twins, under the assumption that such strategy is shaped by anatomical differences. Furthermore, since differences are observed in transitions but not in static targets, this lends support to the hypothesis that acoustic transitions are more likely to reflect individual articulatory variation than target measures.

\subsection{Relationship between articulatory strategy and acoustic variation} \label{articulation}

There is scarce evidence in the literature that could explicitly link specific articulatory strategies to specific acoustic differences in vowels. \citet{noiray2014} show that differences in tongue body height for the production of the /\textipa{I}/-/\textipa{e}/ contrast in American English translate into systematic differences in F1. Several studies observe articulatory differences between males and females, such as increased jaw opening in females \citep{weirich2016, strycharczuk_gender}, and tendency for articulatory undershoot in males \citep{weirich2018individual}, which are linked to acoustic differences, especially larger acoustic vowel spaces in females.
However, the relevant examples are relatively few, whereas a large body of articulatory literature emphasises the relative acoustic invariance produced by diverging articulatory strategies. General plasticity of articulatory strategy is evidenced, for instance, by findings on articulatory compensation triggered by mechanical or auditory perturbation. Speakers whose normal articulation is disrupted, such as by the presence of a bite block, adapt their articulation to approximate their habitual acoustic target \citep{gay1981production, mcfarland1995}. Similar adaptation is observed, in some speakers, when the acoustic feedback received by the speaker is altered experimentally \citep{houde-jordan1998, guenther2016}. These types of adaptation highlight the importance of acoustic invariance over articulatory habit. Classical theories of speech perception postulate a key role of acoustic invariance in phonetics-phonology mappings and establishing a connection between sound and meaning \citep{blumstein1979, lisker1985}, although identifying such regular mappings has often proven challenging \citep{redford2023}. From the sociolinguistic point of view, production of similar acoustic targets serves an indexical role, such as to signal social category membership \citep{eckert2008, foulkesdocherty}.

Acoustic invariance is thought to be the main factor driving the individual differences in articulatory strategy: speakers need to overcome the individual differences in their vocal tract shape and size in order to produce similar sounds. The existence of compensatory mechanisms in speech is evidenced by the relationship between vocal tract morphology and articulatory strategy. \citet{serrurier2024} establish several such correlations. For example, they find a relationship between the horizontal and vertical size of the vocal tract and the primary vector of tongue movement used for achieving formant manipulations. Speakers with relatively larger vocal tracts tend to rely more on anterior-posterior tongue displacement, whereas smaller vocal tracts are associated with more upward-downward displacement. Furthermore, speakers with relatively flatter and more posterior palates tend to achieve F1 manipulations by jaw movement, whereas speakers with more domed and more anterior palates rely more on tongue movement. A similar relationship between palate shape (specifically doming) and jaw movement is also observed by \citet{johnson2023}. 

Compensation for palate shape is well documented in the production of palatal vowels, especially /i/.\footnote{Note that compensation in this case refers to full parasigittal morphology, and not just the midsagittal plane.} \citet{hasegawa2003} demonstrate that the height of the tongue is closely correlated with palate height for palatal vowels /\textipa{i, e, I, E}/, but there is no such correlation in velar, pharyngeal or uvular vowels. Tongue height and palatal vault height were defined as the vertical distance from the line connecting the maxillary gingival margins. Furthermore, the area between the palate and tongue is more stable across speakers in palatal vowels, compared to other vowels.\footnote{A reviewer observes that comparisons of variability between different vowels are complicated by the properties of the vowel inventory in a given language, e.g. variability may be constrained by the presence of multiple vowels in the same area. Another reviewer notes that tongue bracing may add to the relative stability of palatal vowels.} These findings suggest that controlling the degree of vocal tract narrowing is important for achieving the acoustic target in high vowels.

\citet{lammert2013b, lammert2013a} provide further evidence for the systematic relationship between tongue shape in high vowels and the shape of the palate. They identify three parameters that systematically capture individual variation in palate shape: palate concavity (height of the palate apex), anteriority (position of the palate apex) and sharpness (shape at the palate apex). According to modelling in \citet{lammert2013b}, this type of morphological variation has the scope to affect the first three vowel formants substantially. For example, the frequency of F1 increases with greater palate concavity, while the frequency F2 decreases. However, the corresponding acoustic variation is not detected in the speech of individuals with different palate shapes. Instead, speakers adjust the shape of their tongue to compensate for the variation in palate shape; for example, speakers with more concave palates produce more tongue raising in the palatal region. A correlation between palate shape and mean tongue shape is confirmed by \citet{serrurier2023} in a larger sample of 41 speakers. \citet{brunner2009} observe a systematic relationship between articulatory and acoustic variability and palate shape. Specifically, they find that speakers with (cross-sectionally) flat palates show less articulatory variability compared to speakers with domed palates. \citeauthor{brunner2009} propose that this is because the same degree of tongue movement affects the area functions of the vocal tract to a greater degree when the palate is flatter, due to a proportionally smaller vocal tract. Thus, acoustic invariance appears to be driving variation in articulatory strategy through the mechanism of compensation.

\subsection{Present study}
As we have seen, the speaker-specificity of vowel dynamics is an established empirical observation, and it has been attributed to individual differences in articulatory strategy, where such differences may be partially due to compensation for individual differences in vocal tract morphology. On the other hand, existing research on articulatory variation and articulation-acoustics relationship points in a rather different direction, emphasising that articulatory variation largely serves to reduce individual acoustic differences. However, evidence for acoustic compensation is mainly based on static measurements. While compensatory behaviour is undoubtedly an important factor in shaping the speaker's strategy for achieving a specific acoustic target, little is known about its effect on the dynamic properties of the acoustic signal. Dynamics are crucial to consider, given the forensic findings that acoustic dynamics contribute to speaker-specificity, and given the underlying role of articulatory movement. Individual articulatory strategies involve varying degrees of displacement for different articulators. Articulators vary inherently according to their velocity, due to differences in plasticity and mass.  Thus, we can expect that acoustic transitions towards a fixed target might reflect different dynamic properties of the key articulators involved. 

The present study examines the role of articulatory strategy as a factor conditioning vowel dynamics, through a case study of I-diphthongs in Northern-Anglo English  /\textipa{i, eI, aI, oI}/. Note that although we use the /i/ notation for convenience, we consider /\textipa{i}/ to be a diphthong, because it can have a diphthongal quality [\textipa{iI}] and because it behaves like a diphthong structurally \citep{strycharczuk2024_towards}. Our study is based on ultrasound and acoustic data from 36 speakers in the TarDiS corpus \citep{strycharczuk2024_towards, strycharczuk2025_dimensionality}. 

In general, diphthongs offer a compelling case study, because they are inherently dynamic. Furthermore, English diphthongs form a coherent subsystem, which creates an opportunity to examine a wide range of different articulatory displacements, from a small degree of change in [\textipa{iI}] to traversing a large portion of the vocal tract in [\textipa{aI}]. At the same time, all these vowels have an offglide with a similar quality, which allows us to compare whether transitions towards the offglide vary in a similar way across different starting points.

We investigate whether the speaker's formant dynamics in /i/ vowels and diphthongs are systematically affected by their articulatory strategy for /i/ production. As a proxy for this strategy, we use a normalised measure of midsagittal tongue shape in /i/. Specifically, we test whether tongue shape in /i/ is a significant predictor of formant trajectory shape. This is only one of many possible predictors. For example, formant dynamics as well as articulatory strategy are systematically affected by gender\citep{strycharczuk_gender}. Here, we focus on /i/ as a predictor because its relationship to vocal tract morphology is well described \citep{hasegawa2003, lammert2013b}. As a high front vowel, /i/ is systematically affected by the size of the vocal tract, as well as by palate shape (see Section \ref{articulation} above). Thus, we are able to rely on previous literature in interpreting variation in /i/ shape in terms of vocal tract morphology. Another factor that makes /i/ a reasonable predictor is its similarity to the diphthong offglide in other I-diphthongs. Because of the similarity, speakers presumably employ an /i/-like strategy in their transition from the diphthong onglide to the offglide (we test this assumption in Section \ref{pca_results}). Acoustic compensation may be harder to achieve in such cases compared to /i/, in which there is less inherent change and the tongue shape is relatively stable over a longer time window. 

The reason why we use the tongue shape for /i/ to predict formant trajectories and not the actual tongue shape produced in any given vowel at any given time point is to help us tease apart strategy from inter-token variability. We assume that strategy relates to a practised articulatory movement that is relatively stable within speaker (for similar sounds) and different between speakers. However, different speakers may produce different degrees of diphthongisation even when performing the same task, and articulatory variation may also arise within speaker (e.g.  some speakers may produce more vowel reduction as the experiment goes on). The presence of such variation can result in significant correlations between articulatory and acoustic measures, which are, however, relatively trivial (e.g. greater articulatory displacement associated with greater formant displacement). Using a by-speaker mean measure of tongue shape instead helps us isolate the effect of speaker-specific articulatory habit.

In summary, we evaluate the hypothesis that articulatory strategy conditions individuality in vowel dynamics by 
addressing the following research questions:

\begin{enumerate}
    \item What are the different articulatory strategies used by speakers in their production of palatal vowel /i/?
    \item Are these articulatory strategies systematically related to formant dynamics in the I-diphthongs, as produced by the same speakers?
\end{enumerate}

\section{Methods}
\subsection{Corpus}
We present a secondary analysis of a pre-existing corpus, TarDiS. Details of the recording procedure and data processing are described in \citet{kirkham2023_co} and \citet{strycharczuk2025_dimensionality}. Key information relevant to the current study is summarised below.

The corpus comprises time-aligned midsagittal ultrasound and audio recordings from 40 speakers of English from Northern England (Greater Manchester and Lancashire). The ultrasound data were acquired using the EchoB (for 10 speakers) and the Telemed Micro ultrasound speech system (for 30 speakers) from Articulate Instruments Ltd. The sampling rate ranged between 59.5 and 101 frames per second. The median sampling rate was 81.3 frames per second. Ultrafit was used to stabilise the probe \citep{spreafico2018}. In order to image the midsagittal plane, we positioned the transducer on the midline of the body under the participant's chin, and we checked that the tongue contour appeared as single (not parallel) structure in the corresponding ultrasound image.

\subsection{Stimuli}
The corpus includes a full set of stressed vowels of Northern Anglo-English produced within a $b\_d$ environment: \emph{bad, bade, bard, bared, bead, beard, bed, bid, bide, bird, bod, bode, booed, bored, bowed, bud} and \emph{buoyed}. We used all the stimuli in one part of the analysis (Section \ref{pca_results}). For the remainder of the analysis (\ref{gamm_results} and \ref{follow-up}), we focused on the I-diphthong series, represented by \emph{bead}, \emph{bade}, \emph{bide} and \emph{buoyed}. The stimuli were embedded within two types of carrier phrases: \emph{She says X} and \emph{She says X eagerly}. Up to six repetitions (typically five) were produced for each combination of stimulus and carrier phrase by each speaker.

\subsection{Data processing}\label{processing}
The acoustic data were forced aligned using the Montreal Forced Aligner \citep{mcauliffe17}, using the US English MFA language model and the American English dictionary. We acknowledge that there is a difference between the training model and the variety in question, which was mitigated by subsequent manual correction. 

Formant measurements were extracted at every 2ms using FastTrack \citep{barreda2021}, running on Praat version 6.2.14 \citep{praat6214}. These measurements were then normalised using the $\Delta$F method \citep{johnson2020}. The advantage of using this normalisation method is that it takes into account the estimated length of the speaker's vocal tract, which makes it well suited for investigating aspects of articulation-acoustics relationships.

The tongue contour imaged in the ultrasound recordings was automatically labelled using the DeepLabCut (DLC) algorithm \citep{mathis2018}, based on a pre-trained tongue model, described in \citet{wrench2022}. DLC identifies 11 anatomical landmarks in each tongue contour, from the vallecula (most posterior) to the tongue tip (most anterior), which can serve as a basis for speaker normalisation \citep{strycharczuk2025_dimensionality}.

\subsection{Speakers}
The speakers in the TarDiS corpus represent two localities in the North of England: Greater Manchester and Lancashire. Note that, traditionally, the Lancashire dialect and some parts of Greater Manchester may show monophthongisation of selected vowels, particularly the \textsc{face} vowel, which may be realised as [\textipa{e:}] \citep[357]{wells1982.1}. In contrast, \textsc{face} is typically an upgliding diphthong [\textipa{eI}] in the city of Manchester \citep{baranowski2015me} and in General Northern English, the pan-regional standard spoken in the North of England \citep{watt2002, strycharczuk2020}. Since our hypothesis concerns the realisation of I-diphthongs, we excluded speakers with \textsc{face} monophthongisation, whom we identified using the following procedure.

\begin{figure}
    \centering
    \includegraphics[width=1\linewidth]{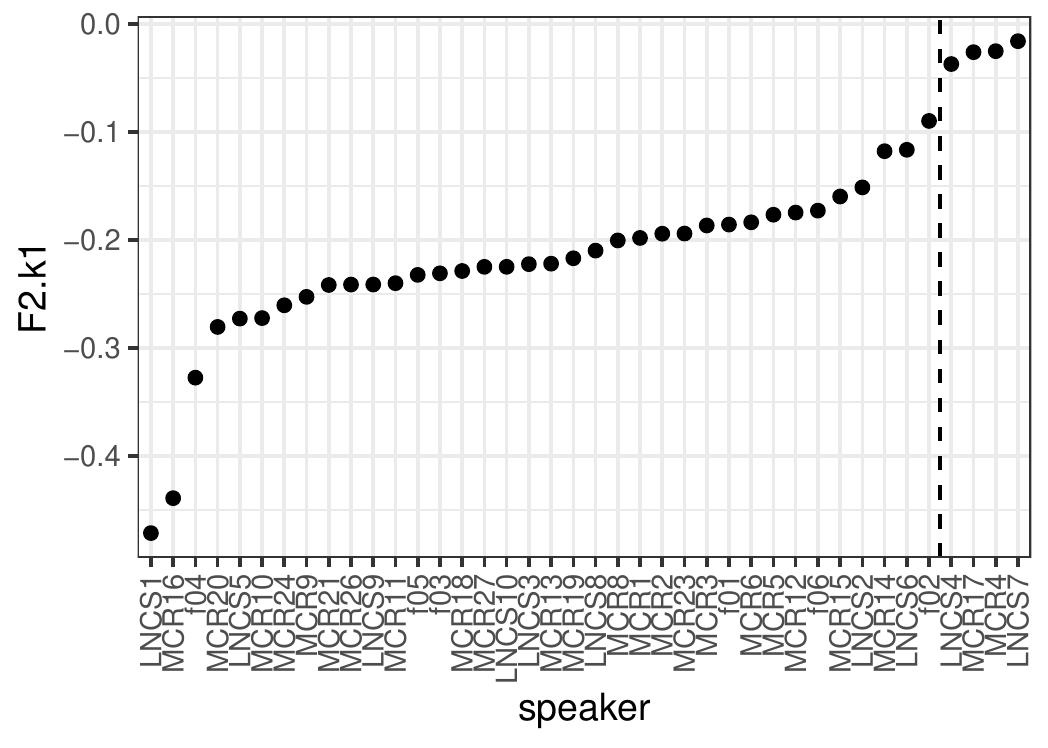}
    \caption{By-speaker mean of the $k_1$ DCT coefficient extracted from the normalised F2 trajectory. Values closer to 0 indicate relatively flatter trajectories (more monophthongal), whereas values further from 0 indicate relatively steeper trajectories (more diphthongal).}
    \label{fig:k1}
\end{figure}

We applied a Discrete cosine transform (DCT) to normalised F2 trajectories \citep{watson1999}. This procedure parametrises the formant trajectory using a number of coefficients. The first three coefficients are respectively related to the mean formant value, the slope of the trajectory, and the trajectory's curvature. Since the second coefficient ($k_1$) corresponds to the slope of the trajectory, it can be taken as a proxy for diphthongisation. If the absolute $k_1$ values are close to zero, this indicates a relatively flat formant trajectory, whereas values further from 0 suggest increasing diphthongisation. Figure \ref{fig:k1} shows the by-speaker mean $k_1$ values for the second formant in our data. These are always negative, which means the diphthong is upgliding. However, we note a range of values, with some of them approaching 0, as would be expected in the presence of variable \textsc{face} monophthongisation. Based on visual inspection of the means, we identify an elbow in the data, which separates four speakers with greatest monophthongisation: LNCS7, MCR4, MCR17 and LNCS4. We excluded these speakers from further analysis, as they cannot be said to produce an /\textipa{I}/ offglide in \emph{bade}. 

The remaining analysis reflects data from 36 speakers. Of these,  19 were female, 15 male, one was non-binary, and one of unknown gender (no response to the gender question). The mean age was 25 (ranging from 18 to 48). The total number of different I-diphthong tokens analysed was 1503.

\subsection{Analysis}
Our primary research question concerns the effect of speaker's tongue shape in /i/ on formant trajectories. To address this question, we need to identify systematic measures of tongue shape that are comparable across speakers. 

In order to derive the relevant measure, we used the method previously described in \citet{lo2025}. First, we applied a Generalised Procrustes Analysis to the tongue contour data. As the input to the analysis we used the tongue contour data representing all individual ultrasound frames corresponding to the vocalic portion in the acoustic signal (e.g. all the available UTI frames for the acoustic portion of /\textipa{i}/ in \emph{bead}). We used the full set of Northern English vowels here to capture the full possible array of vocalic tongue shapes. The analysis performs translation, rotation and scaling to minimise the Procrustes distance between the tongue contours. As such, the analysis normalises for individual differences in tongue size, ultrasound probe placement, probe rotation and movement. The sole feature that differentiates the normalised individual contours is tongue shape. We then performed a principal component analysis (PCA) in the tangent space \citep{kent1992} to identify the main orthogonal dimensions of variation in the tongue shape. This involved projecting the Procrustes-aligned landmark configurations onto the tangent space at the mean shape, which provides a linear approximation to Kendall’s shape space. PCA was then applied to these tangent coordinates so that the resulting components represent interpretable modes of tongue shape variation.

For each speaker, we summarised the tongue shape for /i/ by calculating the mean of the first three principal components (PC1–PC3) across all of their /i/ tokens, averaging over the initial 80\% of each vowel. We excluded the final 20\% of the vowel from the mean calculations, in order to ensure that the mean value is not overly influenced by the following /d/. We then used these by-speaker means as predictors in the statistical modelling of formant trajectories below.

The potential acoustic effects of /i/ tongue shape on diphthong articulation were analysed statistically using Generalised Additive Mixed models (GAMMs; \citealt{wood2017}) with Maximum Likelihood estimation. $\Delta$F-normalised F1 and F2 trajectories were modelled separately for each vowel in the I-Diphthong series (\emph{bead}, \emph{bade}, \emph{bide} and \emph{buoyed}), based on the same following predictor structure:

\begin{itemize}
\item smooth term for duration;
\item smooth term for normalised time;
\item smooth terms for the three /i/ PCs (i-PC1, i-PC2 and i-PC3);
\item tensor product smooth interactions (\texttt{ti} smooths) for the three /i/ PCs  over normalised time;
\item by-speaker random intercept;
\item by-speaker random smooth for normalised time;
\item by-token random smooth for normalised time.
\end{itemize}

Random smooths for time-by-token were included to keep the models maximally conservative \citep{soskuthy2021}. Our main question in analysing the models is whether speakers with different mean /i/ shapes produce significantly different formant trajectories in \emph{bead}, \emph{bade}, \emph{bide} and \emph{buoyed}. 

Significance was established based on model comparison. We ran model comparisons between the full model and a nested model without i-PC1 predictors, a model without i-PC2 predictors, and a model without i-PC3 predictors. An i-PC was considered significant if its inclusion led to a model improvement at $\alpha=.05$.

\section{Results}
\subsection{PCA results}\label{pca_results}
\begin{figure}
    \centering
    \includegraphics[width=1\linewidth]{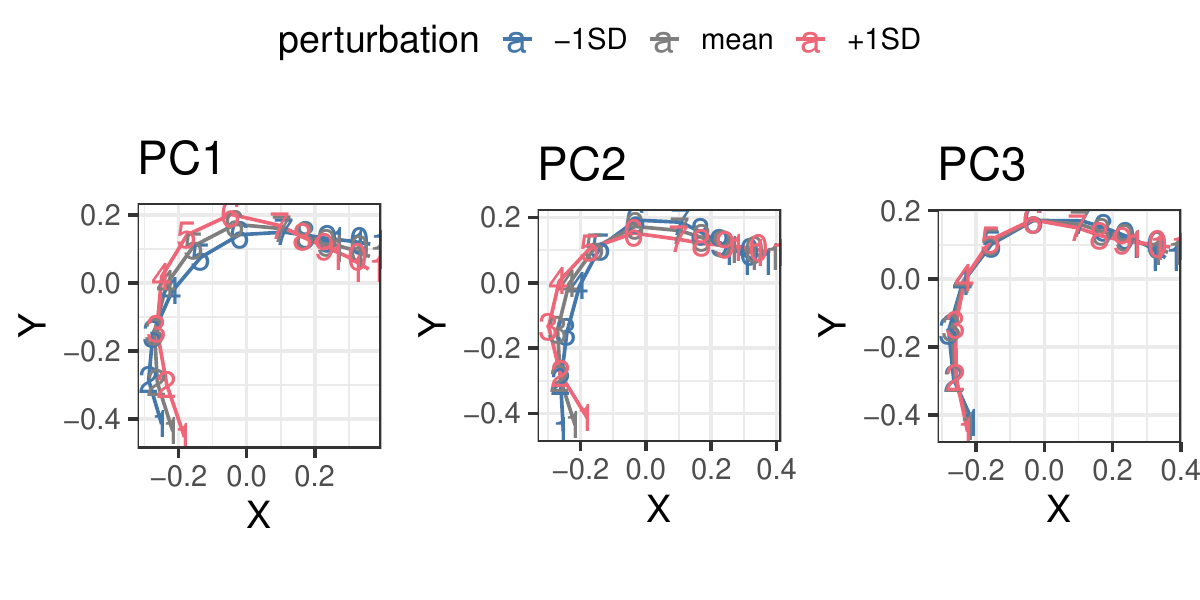}
    \caption{The effect of principal component scores on the variation in tongue shape. The grey line shows the mean tongue shape across all speakers and all vowels. Shapes with relatively higher PC scores (by 1 SD) are plotted in red. Shapes with relatively lower PC scores (by 1 SD) are plotted in blue. The numbers on the tongue contours indicate the location of DLC knots.}
    \label{fig:pca}
\end{figure}

The first three PCs combined captured 79.8\% of variation in tongue shape (recall that this variation is across all vowels). The first three PCs, illustrated in Figure \ref{fig:pca},  capture variation in the degree of tongue concavity/convexity at different locations on the tongue contour, similar to \citet{lo2025}.

PC1 (40.5\% of variance) captures variation in the convexity of tongue shape around the dorsum, centred at DLC knot 5. Higher PC1 scores are associated with more dorsal convexity, whereas lower PC1 scores denote less dorsal convexity and a somewhat flatter tongue shape. PC2 (33.2\% of variance) captures variation in convexity in the anterior part of the tongue dorsum (DLC knots 6 and 7), as well as some tongue root advancement. High PC2 scores are associated with a small degree of concavity in the pre-dorsum, which is accompanied by retraction of the tongue root. In contrast, low PC2 scores are associated with shapes that have arching in the anterior part of the dorsum. Variation in PC3 (6.2\% of variance) is related to variation between convexity (low PC3 scores) and concavity (high PC3 scores) in the tongue mid, centred at DLC knot 8. PCA predictions for tongue variation in /i/ are plotted in Figure \ref{fig:fig3}. While the general interpretation of the PCs remains the same, the figure visualises the range of variation (within 1 SD) associated with different articulatory strategies for /i/ in our data.

Table \ref{tab:sumstats} summarises the means and standard deviations of the PC scores in the dataset across all vowels and in /i/, and reports the standard errors of the mean. These summary statistics are calculated over the initial 80\% of the vowel, in order to exclude tongue shapes that are likely strongly influenced by the following /d/. Overall variance is greater across vowels than within /i/, which is expected, since /i/ variance spans only inter-speaker and token variation, whereas across vowels, variation is induced by both vowel and speaker. However, even within /i/ we find substantial variation in PC scores, suggesting considerable speaker variation in tongue shape. On average, /i/ has higher PC1 than the grand mean PC1 value. This is expected given the interpretation that PC1 is a correlate of increased doming at tongue dorsum -- /i/ being a high vowel has a considerable degree of dorsal doming. Mean PC2 score in /i/ is negative, that is, lower compared to the grand mean PC2 score. This aligns with the interpretation of low PC2 values representing a combination of tongue doming and anteriority (more anterior location of the tongue constriction). In contrast to PC1 and PC2, the central tendency of PC3 is similar in /i/ as across all vowels, suggesting that the tongue shape information carried by PC3 is similar in /i/ and in the speaker's overall mean tongue shape. The SEM values for /i/ are larger than for the whole dataset, reflective of the fact that the /i/ tokens represent a subset of the data.

\begin{table}

\caption{\label{tab:sumstats}Summary statistics for PC scores across all vowels and within /\textipa{i}/}
\centering
\begin{tabular}{l|rrr|rrr}
  \hline
 \multicolumn{1}{c|}{} & \multicolumn{3}{c|}{all vowels} & \multicolumn{3}{c}{/\textipa{i}/} \\
PC & mean & SD & SEM & mean & SD & SEM  \\ 
  \hline
PC1 & 0.008 & 0.089 & 0.0003 & 0.021 & 0.078 & 0.0011 \\ 
  PC2 & 0.004 & 0.082 & 0.0003 & -0.101 & 0.061 & 0.0008 \\ 
  PC3 & -0.003 & 0.035 & 0.0001 & 0.001 & 0.034 & 0.0005 \\ 
   \hline
\end{tabular}
\end{table}

We can also use the PC scores to verify one of the key assumptions of this study, namely that the mean tongue shape in /i/ is closely related to the shape of I-diphthong offglides. Table \ref{tab:corr} summarises the results of a series of Pearson's correlation tests, examining the correlation between by-speaker mean PC scores for /i/ and by-speaker mean scores for other I-diphthong offglides. The mean PC scores for /i/ are calculated over the initial 80\% of the vowel, as in Table \ref{tab:sumstats}.  The offglide was defined as the ultrasound frame closest to the 85\% of vowel duration. This time point was selected based on our previous research into diphthong timing in the same dataset \citep{strycharczuk2024_towards}. As we can see from the table, the strength of the correlation was very high or high for PC1 and PC2 (correlation coefficients ranging from 0.85 to 0.95). In comparison, PC3 shows somewhat weaker, though still quite substantial, correlations between /i/ and diphthong offglides: 0.77 for the offglide of \emph{bade}, 0.65 for the offglide of \emph{bide}, and 0.67 for the offglide of \emph{buoyed}. From this, we can conclude that the mean tongue shape in /i/ is strongly isomorphic with the tongue shape of the same speaker's diphthong offglide, especially in terms of palate doming and anteriority. Having verified this assumption, let us consider whether the variation in /i/ tongue shape affects formant trajectories in I-diphthongs.

\begin{table}[ht]
\centering
\caption{Pearson's correlations between by-speaker mean PC for each I-diphthong offglide and the corresponding i-PC (all $p<.001$)}
\label{tab:corr}
\begin{tabular}{lrrr}
  \hline
PC & \emph{bade} & \emph{bide} & \emph{buoyed} \\ 
  \hline
PC1 & 0.93 & 0.88 & 0.89 \\ 
  PC2 & 0.95 & 0.85 & 0.88 \\ 
  PC3 & 0.77 & 0.65 & 0.67 \\ 
   \hline
\end{tabular}
\end{table}

\subsection{GAMM results} \label{gamm_results}
Table \ref{tab:psum} summarises the results of ML comparison testing the significance of individual i-PCs as predictors of time-varying F1 and F2. An i-PC was considered significant if its inclusion in the model significantly improved the model fit. 

As we can see from the table, i-PC1 significantly predicted the F1 trajectory shape for \emph{bade, bide}, and \emph{buoyed} ($p<.001$), but not for \emph{bead} ($p=0.950$). For F2 trajectory shape, i-PC1 was significant only for \emph{buoyed} and \emph{bead} ($p<.001$), while the effects for \emph{bade} and \emph{bide} were not significant. i-PC2 significantly predicted F1 for all four vowels. i-PC3 showed significant effects on F1 trajectory for all vowels, including bead ($p=0.044$), but for F2 it was significant only for \emph{buoyed} and \emph{bead} ($p<.001$).

\begin{table}[ht]
\centering
\caption{Summary of $p$-values returned by model comparisons between a full GAMM model of formant trajectory and a nested model without a specific i-PC.}
\label{tab:psum}
\centering
\begin{tabular}{llrrrr}
  \hline
predictor & DV & bade & bide & buoyed & bead \\ 
  \hline
i-PC1 & F1 & $<$.001 & $<$.001 & $<$.001 & 0.950 \\ 
  i-PC1 & F2 & 0.186 & 0.083 & $<$.001 & $<$.001 \\ 
  i-PC2 & F1 & $<$.001 & $<$.001 & $<$.001 & 0.003 \\ 
  i-PC2 & F2 & 0.010 & 0.021 & $<$.001 & 0.005 \\ 
  i-PC3 & F1 & $<$.001 & $<$.001 & $<$.001 & 0.044 \\ 
  i-PC3 & F2 & 0.529 & 0.576 & $<$.001 & $<$.001 \\ 
   \hline
\end{tabular}
\end{table}

\subsection{Follow-up exploratory analysis} \label{follow-up}
\begin{figure}
    \centering
    \includegraphics[width=0.5\linewidth]{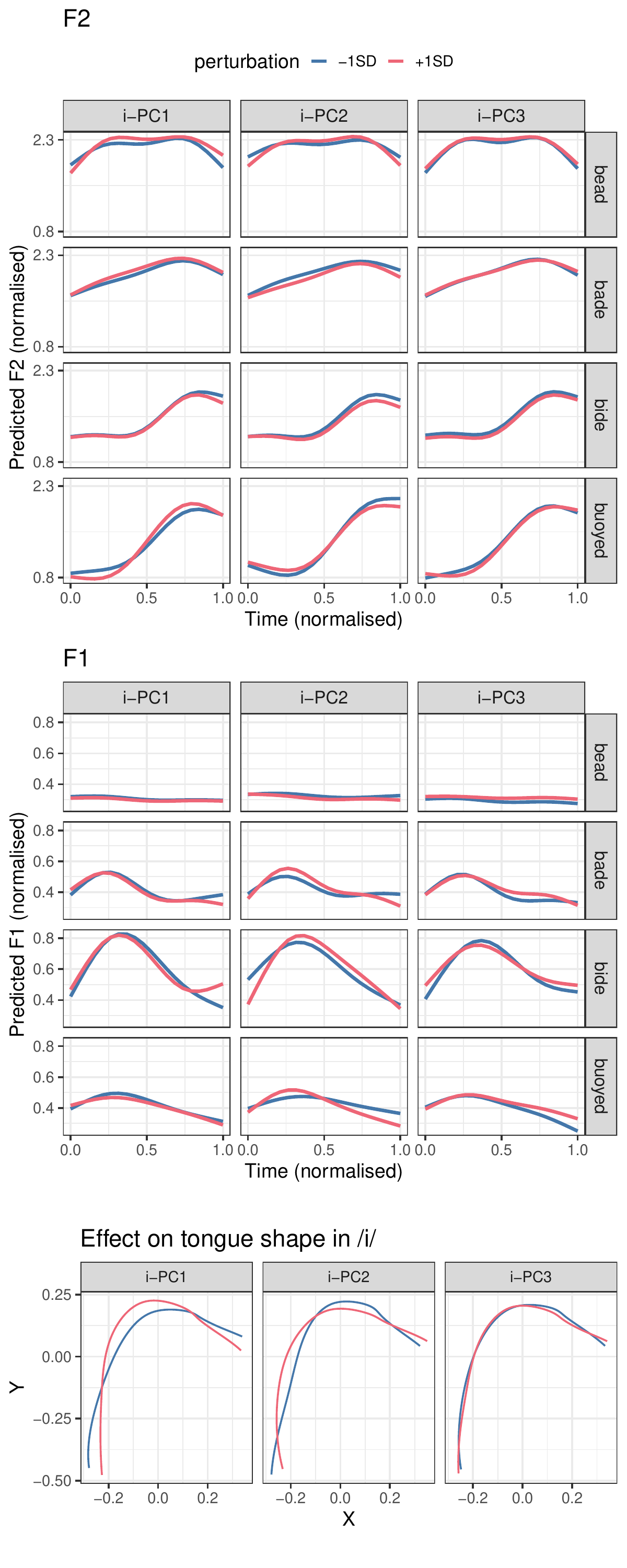}
    \caption{Top: GAMM predictions for normalised F1 and F2 trajectories, depending on tongue shape (i-PC value) and vowel.  Bottom: Visualisation of predicted tongue shape for the speakers with high (red) and low (blue) i-PC values.}
    \label{fig:fig3}
\end{figure}

In order to understand the nature of the i-PC effects summarised in Table \ref{tab:psum}, we visualised the model predictions, comparing speakers with high and low i-PC values, where `high' and `low' are defined as being $\pm$1SD from the mean i-PC value. Figure \ref{fig:fig3} illustrates predicted formant trajectories for normalised F1 and F2, depending on tongue shape and vowel.  The predicted formant trajectories for speakers with high (1SD above the mean) and low i-PC values (1SD below the mean) are plotted in red and blue respectively. To aid interpretation, the predicted tongue shapes in /i/, depending on PC perturbation, are included in the bottom panel of the same figure.  

While the effect of i-PC1 on F1 trajectories in \emph{bade}, \emph{bide} and \emph{buoyed} was significant (see Table \ref{tab:psum}), it is very small in scope. In comparison, the effect of i-PC1 on the F2 trajectory in \emph{buoyed} is more readily observable. Nevertheless, across these four cases, we can generalise that the acoustic shift from onglide to offglide starts earlier in speakers with high i-PC1 (speakers with more dorsal doming).

Variation in i-PC2 was significantly associated with trajectory shape for all vowels and trajectories, according to Table \ref{tab:psum}. According to Figure \ref{fig:fig3}, speakers with low i-PC2 scores (more domed tongue shape in /i/ with more anterior constriction) generally produce  earlier and faster formant transitions. The differences are larger for F1 than for F2. The F1 trajectory in \emph{buoyed} is somewhat exceptional here, in that speakers with low i-PC2 maintain a relatively more stable F1 over the whole duration of the vowel, compared to speakers with high i-PC2 scores, such that the offglide is slightly acoustically undershot.

Variation in i-PC3 mainly affects the offglide F1. For \emph{bade}, \emph{bide} and \emph{buoyed}, speakers with high i-PC3 values achieve a slightly higher minimum F1 in the offglide, compared to speakers with low i-PC3. The effect of i-PC3 on F2 trajectories is overall very small, although it was significant for \emph{bead} and \emph{buoyed}.

 In order to further illuminate the observed differences, we conducted an analysis of vowel duration. Our rationale for analysing duration is that we observe differences in the relative timing of formant trajectories in normalised time. It is therefore important to ensure that the differences are not an artifact of time normalisation and fundamentally caused by systematic differences in vowel duration. To test the effect of i-PCs on duration, we fitted a linear mixed effects regression model with vowel duration as dependent variable, and i-PCs 1, 2 and 3 as predictors. Vowel (/\textipa{i}/, /\textipa{eI}/, /\textipa{aI}/ and /\textipa{oI}/) and speaker were included as random intercepts. $P$-values for coefficients were obtained using Satterthwaite's approximation \citep{lmertest}. The model summary is in Table \ref{tab:durlmer}. As we can see from the table, none of the i-PCs was significant. This suggests that the differences we identify in formant dynamics conditioned by i-PCs are not related to any underlying differences in vowel duration.

\begin{table}[ht]
\centering
\caption{Model summary for a linear mixed effects model predicting vowel duration.}
\label{tab:durlmer}

\begin{tabular}{lrrrr}
  \hline
term & $\beta$ & SE & $t$ & $p$ \\ 
  \hline
(Intercept) & 0.27 & 0.03 & 10.49 & $<$.001 \\ 
  bead.PC1 & 0.02 & 0.15 & 0.17 & 0.868 \\ 
  bead.PC2 & 0.09 & 0.20 & 0.45 & 0.658 \\ 
  bead.PC3 & -0.47 & 0.43 & -1.08 & 0.287 \\ 
   \hline
\end{tabular}
\end{table}

Finally, we conducted an analysis of changes in tongue shape and articulatory displacement, as a function of the /i/-PCs. Our aim was to understand the nature of movement produced in diphthongs by speakers with different tongue shapes in /i/. In the articulatory analysis, we used a range of articulatory variables, measured dynamically, as dependent variables in GAMMs. The variables were: dynamic change in PC1, PC2 and PC3, fronting of the tongue root (horizontal displacement of DLC knot 4), fronting of the tongue dorsum (horizontal displacement of DLC knot 5), raising of the posterior and anterior parts of the tongue dorsum (vertical displacement of DLC knots 5 and 7), raising of the tongue mid (knot 8), and raising of the tongue blade (knot 10). The choice of these variables was informed by our previous research on parametrising tongue shape variation in vowels \citep{strycharczuk2025_dimensionality}. All the displacement variables were rotated to the speaker's occlusal plane and centred within speaker. Note that the tongue coordinates were not corrected relative to any hard structures in the vocal tract, such as the palate. Consequently, it is possible that some of the height measures are overestimated in low vowels, and somewhat underestimated in high vowels. We do not consider this issue to be severe, based on good overall agreement between DLC-based measures of tongue position and height, compared to corresponding Electromagnetic Articulometry results \citep{kirkham2025}.

The predictor structure in the articulatory models was identical to the acoustic models reported above, except we used Fast Restricted Maximum Likelihood estimation. The full details of all the models are reported in the online repository. Here, we report selected key results, focusing on the articulatory variables that we expect to co-vary systematically with tongue shape in /\textipa{i}/, based on our interpretation of the PCA presented in Section \ref{pca_results}. 

\begin{figure}
    \centering
    \includegraphics[width=1\linewidth]{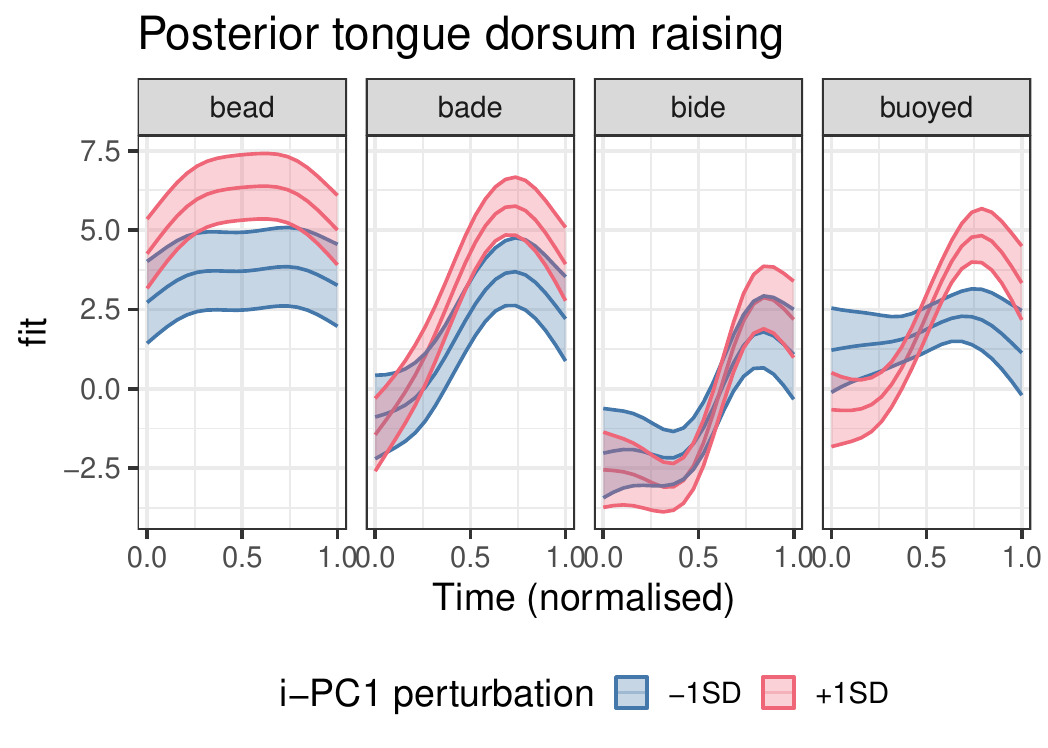}
    \caption{GAMM predictions for patterns of raising of the posterior part of the dorsum (DLC knot 5) in speakers with high and low values of i-PC1}
    \label{fig:fig4}
\end{figure}

Figure \ref{fig:fig4} shows predicted patterns of tongue dorsum raising in speakers with high and low i-PC1. Speakers with relatively higher i-PC1 scores show considerably more dorsum raising in \emph{bead} and in \emph{bade} and \emph{buoyed} offglides, consistent with greater doming of the tongue dorsum as captured by PC1. Additionally, the tongue dorsum raising is more rapid in \emph{bade} and \emph{buoyed}. 

\begin{figure}
    \centering
    \includegraphics[width=1\linewidth]{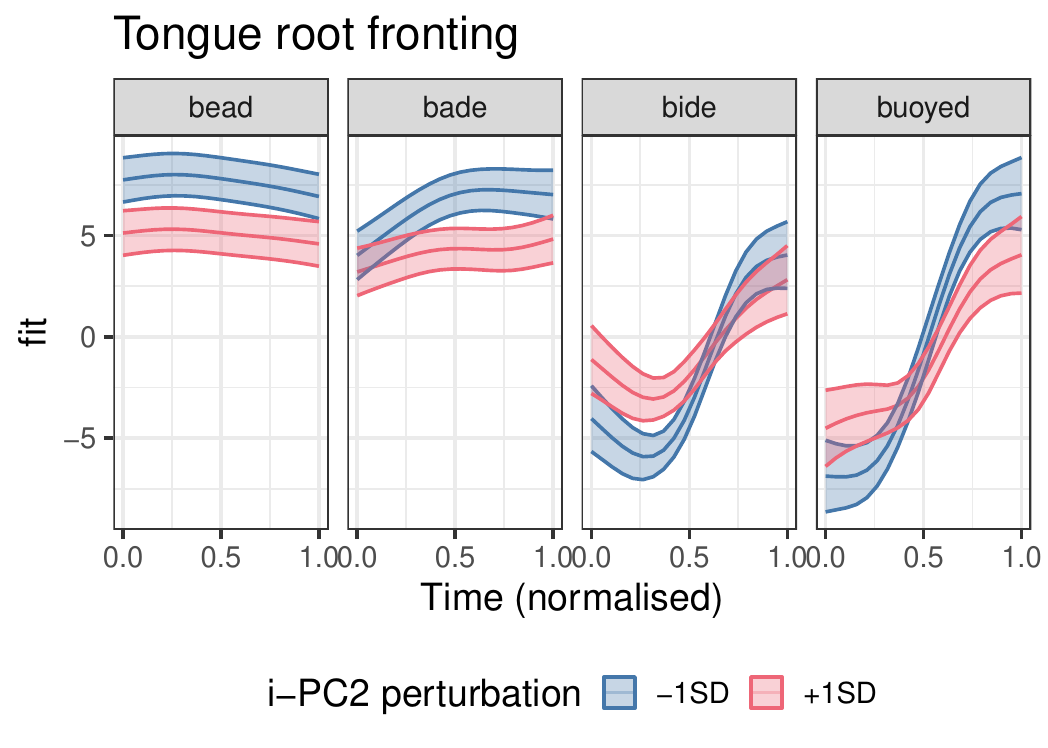}
    \caption{GAMM predictions for patterns of tongue root fronting in I-diphthongs (DLC knot 4) in speakers with high and low values of i-PC2}
    \label{fig:fig5}
\end{figure}

Figure \ref{fig:fig5} illustrates the predicted trajectory of tongue root fronting in I-diphthongs, depending on variation in i-PC2. Low i-PC2 scores are associated with a more domed tongue shape and a more anterior constriction location (Figure \ref{fig:fig3}). In accordance with this, speakers with relatively lower PC2 scores have more tongue root fronting in \emph{bead} and in I-diphthong offglides, compared to speakers with high i-PC2 scores. Greater tongue-root advancement may be necessary for speakers with highly domed palates, allowing those speakers to raise the dorsum closer to the palate. Low i-PC2 scores are also associated with more inherent change in tongue root fronting in \emph{bide} and \emph{buoyed}. This suggests a greater horizontal articulatory range in speakers with low i-PC2 scores. 

Compared to i-PCs 1 and 2, variation in i-PC3 is not linked to robust differences in articulatory displacement. We return to this point in Section \ref{discussion} below.

\section{Discussion} \label{discussion}

We have presented evidence that speakers who share specific tongue shape features in their /i/ production also share some features of their acoustic formant trajectories in selected I-diphthongs. This result provides support to the hypothesis that articulatory strategy has a systematic influence on formant transitions. More importantly, our findings illuminate the nature of this influence, by linking specific types of articulatory strategy for palatal vowel production with specific acoustic dynamic changes. In general, we find that speakers who have a more anterior constriction, or more domed tongue shape, in /i/ produce earlier formant transitions between onglide and offglide in selected I-diphthongs. 

The relationship between tongue shape and formant dynamics can be explained by the properties of articulatory movement associated with different tongue shapes and their interaction with palatal shape. Some tongue shapes involve more articulatory displacement than others: more domed tongue shape is associated with greater tongue dorsum raising (Figure \ref{fig:fig4}) and producing a constriction in the front section of the tongue involves more tongue root fronting (Figure \ref{fig:fig5}). This means that speakers with high PC1 and speakers with low PC2 have to traverse a larger distance within the same time window. Note that we find no differences in vowel duration depending on tongue shape (Table \ref{tab:durlmer}). Given the need to cover greater distance and a stable time interval, these speakers need to produce higher velocity, as well as faster movement, at the onset of the gesture. This is consistent with the predictions of a linear harmonic oscillator model for tract variable motion \citep{saltzman1989}. Specifically, under the assumption of identical activation durations, damping and stiffness parameters, movements across different distances will converge to the target at the same time point, with larger movements having a much steeper velocity slope and higher peak velocity. The differences observed in our data can therefore be understood as a simple consequence of variable movement distances in differently sized vocal tracts, and potentially with tongues of differing volume, with similar timing of target achievement being privileged across speakers.

Having discussed why the specific articulatory strategies condition the observed changes in formant dynamics, let us also consider why we find the specific strategies in the first place. Variation in i-PC1 and i-PC2 aligns closely with previous findings on articulatory strategy and vocal tract morphology \citep{lammert2013b, serrurier2019, serrurier2023, serrurier2024}. Variation in i-PC1 is characterised by different degrees of dorsal doming. Increased doming of the tongue dorsum involves more dorsal raising. Additionally, more domed /i/ shapes are associated with a greater vertical range of dorsal displacement. This type of articulatory strategy is possibly conditioned by greater vocal tract size in the vertical dimension, and possibly also by greater palate doming. The tongue root differences between speakers with low and high i-PC2 are also most likely driven by articulatory compensation. Previous research suggests that increased range of tongue root movement and greater palate doming are characteristic of relatively longer vocal tracts \citep{serrurier2024}.

While the proposed mechanism can account for the effect of i-PC1 (dorsal doming) and i-PC2 (anterior doming and tongue root advancement), the effect of i-PC3 on formant transitions in \emph{bade} is more challenging to account for. Unlike i-PC1 and i-PC2, variation in i-PC3 does not seem to be linked to major differences in articulatory displacement, and so it probably is not conditioned by differences in vocal tract size or shape. We hypothesise that speakers with high i-PC3 have a greater than average tendency for anticipatory coarticulation between the vowel and the following consonant, in this case /d/. The slight concavity in the anterior part of the tongue for these speakers (Figure \ref{fig:fig3}) could be caused by tongue tip raising in anticipation of the following alveolar closure. This interpretation is consistent with the observation that speakers with relatively high i-PC3 tend to produce higher F1 in their I-diphthong offglides, suggesting a small degree of offglide undershoot, which could be consistent with increased coarticulation. This explanation, however, is only tentative, and it invites follow-up research with more direct evidence on palate shape.

We observe that \emph{bead} consistently shows less robust acoustic differences related to tongue shape, compared to other I-diphthongs (Figure \ref{fig:fig3}). This aligns with earlier forensic findings that speaker-specificity of formant trajectories interacts with vowel, such that greater inherent vowel change (more diphthongisation) is associated with greater speaker-specificity of formant transitions \citep{heeren2020}. Both sets of results suggest that the acoustic effect of articulatory strategy is more readily observable where there is larger articulatory displacement. While speakers are generally able to compensate for differences in their vocal tract shape, instances of large articulatory displacement can occasionally reveal underlying individual strategy variation in the resulting acoustics. 

\section{Conclusion}
Our results provide empirical support for the hypothesis that vowel dynamics can systematically differ depending on the characteristics of articulatory strategy in palatal vowel production, which itself is likely a reflection of vocal tract constraints. The link between the nature of articulation and acoustic vowel dynamics is mediated by systematic properties of articulatory movement. Specifically, greater articulatory displacement is associated with somewhat earlier and more rapid formant transitions. We have argued that these effects arise from the dynamic properties of the gestural organisation of speech, which condition a faster velocity rise when greater displacement is required within a fixed time window. This mechanism explains the link between vocal tract constraints on the one hand and individual differences in vowel dynamics on the other, also highlighting the limits of articulatory compensation: while articulatory variation is largely conditioned by compensation in the first place, some aspects of the relevant variation are acoustically recoverable. As such, our study bridges earlier forensic and articulatory findings, and advances the conceptual understanding of the sources of individuality in speech. It also calls for further investigation into the generalisability of the relationship between greater articulatory displacement and formant trajectory shape.

Our study is informed by data from speakers of English from a specific locality. We have no reason to suspect that the relationship between the articulatory strategy and formant dynamics that we observe is specific to English, which is why we explain it in terms of universal principles of the articulation-acoustics relationship. 
However, we cannot rule out the possibility that our results are influenced, for example, by language-specific articulatory setting \citep{laver1980}. Our conclusion that speaker anatomy can influence formant dynamics via articulatory strategy would be greatly strengthened by evidence from other languages and/or from studies focusing on other aspects of anatomy and articulatory strategy.

\begin{acknowledgments}
We thank the JASA editor and reviewers for their feedback on the earlier versions of this manuscript. This research was supported by an AHRC research grant awarded to PS and SK (AH/S011900/1), a British Academy Mid-Career Fellowship awarded to PS (MFSS24\textbackslash240076), an AHRC Fellowship awarded to SK (AH/Y002822/1), and a Royal Society grant also supported by the Leverhulme Trust (APX\textbackslash R1\textbackslash251102) awarded to S.K. We would like to acknowledge the assistance given by Research IT and the use of the Computational Shared Facility at The University of Manchester.

\end{acknowledgments}

\section*{Author declarations}
The authors have no conflicts to disclose.

\section*{Ethics approval statement}
The study was exempt from ethical approval, as it constitutes secondary data analysis. The data collection for the original corpus received ethical approval and the participants provided informed consent, as reported in Strycharczuk et al. (2024, 2025).

\section*{Data availability statement} 

The data and code reported in this paper are publicly available as an OSF repository at \url{https://osf.io/xtp6q/}.


\begin{thebibliography}{58}
\def\enquote#1{``#1,''}
\def\plainquote#1{``#1''}
\expandafter\ifx\csname natexlab\endcsname\relax\def\natexlab#1{#1}\fi
\providecommand{\dourl}[1]{\href{http://#1}{\nolinkurl{#1}}}
\providecommand{\bibinfo}[2]{#2}
\providecommand{\noopsort}[1]{}
\providecommand{\switchargs}[2]{#2#1}
  \def\eatspace #1{#1}

\bibitem[{Baranowski and Turton(2015)}]{baranowski2015me}
\bibinfo{author}{Baranowski, M.},  and \bibinfo{author}{Turton, D.}
  (\textbf{\bibinfo{year}{2015}}). \enquote{\bibinfo{title}{Manchester
  {E}nglish}} in \emph{\bibinfo{booktitle}{Researching Northern Englishes}},
  edited by \bibinfo{editor}{R.~Hickey} (\bibinfo{publisher}{John Benjamins},
  \bibinfo{address}{Amsterdam and Philadelphia}), pp.
  \bibinfo{pages}{293--316}.

\bibitem[{Barreda(2021)}]{barreda2021}
\bibinfo{author}{Barreda, S.} (\textbf{\bibinfo{year}{2021}}).
  \enquote{\bibinfo{title}{Fast track: fast (nearly) automatic formant-tracking
  using {P}raat}} \bibinfo{journal}{Linguistics Vanguard} \textbf{7}(1),
  \bibinfo{pages}{20200051}.

\bibitem[{Blumstein and Stevens(1979)}]{blumstein1979}
\bibinfo{author}{Blumstein, S.~E.},  and \bibinfo{author}{Stevens, K.~N.}
  (\textbf{\bibinfo{year}{1979}}). \enquote{\bibinfo{title}{Acoustic invariance
  in speech production: Evidence from measurements of the spectral
  characteristics of stop consonants}} \bibinfo{journal}{J. Acoust. Soc. Am.}
  \textbf{66}(4), \bibinfo{pages}{1001--1017}.

\bibitem[{Boersma and Weenink(2022)}]{praat6214}
\bibinfo{author}{Boersma, P.},  and \bibinfo{author}{Weenink, D.}
  (\textbf{\bibinfo{year}{2022}}). \plainquote{\bibinfo{title}{Praat: doing
  phonetics by computer [{C}omputer programme]}} \dourl{http://www.praat.org/},
  \bibinfo{note}{{V}ersion 6.2.14}.

\bibitem[{Brunner \emph{et~al.}(2009)Brunner, Fuchs, and Perrier}]{brunner2009}
\bibinfo{author}{Brunner, J.}, \bibinfo{author}{Fuchs, S.},  and
  \bibinfo{author}{Perrier, P.} (\textbf{\bibinfo{year}{2009}}).
  \enquote{\bibinfo{title}{On the relationship between palate shape and
  articulatory behavior}} \bibinfo{journal}{J. Acoust. Soc. Am.}
  \textbf{125}(6), \bibinfo{pages}{3936--3949}.

\bibitem[{Eckert(2008)}]{eckert2008}
\bibinfo{author}{Eckert, P.} (\textbf{\bibinfo{year}{2008}}).
  \enquote{\bibinfo{title}{Variation and the indexical field}}
  \bibinfo{journal}{J. Sociolinguistics} \textbf{12}(4),
  \bibinfo{pages}{453--476}.

\bibitem[{Foulkes and Docherty(2006)}]{foulkesdocherty}
\bibinfo{author}{Foulkes, P.},  and \bibinfo{author}{Docherty, G.}
  (\textbf{\bibinfo{year}{2006}}). \enquote{\bibinfo{title}{The social life of
  phonetics and phonology}} \bibinfo{journal}{J. Phonetics} \textbf{34},
  \bibinfo{pages}{409--438}.

\bibitem[{Gay \emph{et~al.}(1981)Gay, Lindblom, and Lubker}]{gay1981production}
\bibinfo{author}{Gay, T.}, \bibinfo{author}{Lindblom, B.},  and
  \bibinfo{author}{Lubker, J.} (\textbf{\bibinfo{year}{1981}}).
  \enquote{\bibinfo{title}{Production of bite-block vowels: {A}coustic
  equivalence by selective compensation}} \bibinfo{journal}{J. Acoust. Soc.
  Am.} \textbf{69}(3), \bibinfo{pages}{802--810}.

\bibitem[{Guenther(2016)}]{guenther2016}
\bibinfo{author}{Guenther, F.~H.} (\textbf{\bibinfo{year}{2016}}).
  \emph{\bibinfo{title}{Neural Control of Speech}} (\bibinfo{publisher}{The MIT
  Press}, \bibinfo{address}{Cambridge, MA}).

\bibitem[{Hasegawa-Johnson \emph{et~al.}(2003)Hasegawa-Johnson, Pizza, Alwan,
  Cha, and Haker}]{hasegawa2003}
\bibinfo{author}{Hasegawa-Johnson, M.}, \bibinfo{author}{Pizza, S.},
  \bibinfo{author}{Alwan, A.}, \bibinfo{author}{Cha, J.~S.},  and
  \bibinfo{author}{Haker, K.} (\textbf{\bibinfo{year}{2003}}).
  \enquote{\bibinfo{title}{Vowel category dependence of the relationship
  between palate height, tongue height, and oral area}} \bibinfo{journal}{J.
  Speech Lang. Hear. Res.} \textbf{46}(3), \bibinfo{pages}{738--753}.

\bibitem[{Heeren(2020)}]{heeren2020}
\bibinfo{author}{Heeren, W.} (\textbf{\bibinfo{year}{2020}}).
  \enquote{\bibinfo{title}{The contribution of dynamic versus static formant
  information in conversational speech}} \bibinfo{journal}{Int. J. Speech Lang.
  Law} \textbf{27}(1), \bibinfo{pages}{75--98}.

\bibitem[{Houde and Jordan(1998)}]{houde-jordan1998}
\bibinfo{author}{Houde, J.},  and \bibinfo{author}{Jordan, M.~I.}
  (\textbf{\bibinfo{year}{1998}}). \enquote{\bibinfo{title}{Sensorimotor
  adaptation in speech production}} \bibinfo{journal}{Science}
  \textbf{279}(5354), \bibinfo{pages}{1213--1216}.

\bibitem[{Hughes \emph{et~al.}(2016)Hughes, Wood, and Foulkes}]{hughes2016}
\bibinfo{author}{Hughes, V.}, \bibinfo{author}{Wood, S.},  and
  \bibinfo{author}{Foulkes, P.} (\textbf{\bibinfo{year}{2016}}).
  \enquote{\bibinfo{title}{Strength of forensic voice comparison evidence from
  the acoustics of filled pauses}} \bibinfo{journal}{Int. J. Speech Lang. Law}
  \textbf{23}(1), \bibinfo{pages}{99--132}.

\bibitem[{Johnson(2020)}]{johnson2020}
\bibinfo{author}{Johnson, K.} (\textbf{\bibinfo{year}{2020}}).
  \enquote{\bibinfo{title}{The {$\Delta$F} method of vocal tract length
  normalization for vowels}} \bibinfo{journal}{Lab. Phonol.} \textbf{11}(1),
  \bibinfo{pages}{10}.

\bibitem[{Johnson(2023)}]{johnson2023}
\bibinfo{author}{Johnson, K.} (\textbf{\bibinfo{year}{2023}}).
  \enquote{\bibinfo{title}{Individual differences in speech production: What is
  ``phonetic substance''?}} in \emph{\bibinfo{booktitle}{Proc. 20th Inter.
  Congr. {P}honetic Sci.}}, edited by \bibinfo{editor}{R.~Skarnitzl} and
  \bibinfo{editor}{J.~Vol\'{i}n}, \bibinfo{publisher}{International Phonetic
  Association}, pp. \bibinfo{pages}{1102--1106}.

\bibitem[{Johnson \emph{et~al.}(1993)Johnson, Ladefoged, and
  Lindau}]{johnson1993}
\bibinfo{author}{Johnson, K.}, \bibinfo{author}{Ladefoged, P.},  and
  \bibinfo{author}{Lindau, M.} (\textbf{\bibinfo{year}{1993}}).
  \enquote{\bibinfo{title}{Individual differences in vowel production}}
  \bibinfo{journal}{J. Acoust. Soc. Am.} \textbf{94}(2),
  \bibinfo{pages}{701--714}.

\bibitem[{Kent(1992)}]{kent1992}
\bibinfo{author}{Kent, J.~T.} (\textbf{\bibinfo{year}{1992}}).
  \enquote{\bibinfo{title}{New directions in shape analysis}} in
  \emph{\bibinfo{booktitle}{The Art of Statistical Science}}, edited by
  \bibinfo{editor}{K.~V. Mardia} (\bibinfo{publisher}{Wiley},
  \bibinfo{address}{New York}), pp. \bibinfo{pages}{115--127}.

\bibitem[{Kirchhoff \emph{et~al.}(2002)Kirchhoff, Fink, and
  Sagerer}]{kirchhoff2002}
\bibinfo{author}{Kirchhoff, K.}, \bibinfo{author}{Fink, G.~A.},  and
  \bibinfo{author}{Sagerer, G.} (\textbf{\bibinfo{year}{2002}}).
  \enquote{\bibinfo{title}{Combining acoustic and articulatory feature
  information for robust speech recognition}} \bibinfo{journal}{Speech Commun.}
  \textbf{37}(3--4), \bibinfo{pages}{303--319}.

\bibitem[{Kirkham and Strycharczuk(2025)}]{kirkham2025}
\bibinfo{author}{Kirkham, S.},  and \bibinfo{author}{Strycharczuk, P.}
  (\textbf{\bibinfo{year}{2025}}). \enquote{\bibinfo{title}{Dynamical model
  parameters from ultrasound tongue kinematics}} \bibinfo{journal}{JASA Express
  Lett.} \textbf{5}(11), \bibinfo{pages}{115201}.

\bibitem[{Kirkham \emph{et~al.}(2023)Kirkham, Strycharczuk, Gorman, Nagamine,
  and Wrench}]{kirkham2023_co}
\bibinfo{author}{Kirkham, S.}, \bibinfo{author}{Strycharczuk, P.},
  \bibinfo{author}{Gorman, E.}, \bibinfo{author}{Nagamine, T.},  and
  \bibinfo{author}{Wrench, A.} (\textbf{\bibinfo{year}{2023}}).
  \enquote{\bibinfo{title}{Co-registration of simultaneous high speed
  ultrasound and electromagnetic articulography for speech production
  research}} in \emph{\bibinfo{booktitle}{Proc. 20th {I}nter. {C}ongr.
  {P}honetic {S}ci.}}, edited by \bibinfo{editor}{R.~Skarnitzl} and
  \bibinfo{editor}{J.~Vol\'{i}n}, \bibinfo{publisher}{International Phonetic
  Association}, pp. \bibinfo{pages}{918--922}.

\bibitem[{Kuznetsova \emph{et~al.}(2017)Kuznetsova, Brockhoff, and
  Christensen}]{lmertest}
\bibinfo{author}{Kuznetsova, A.}, \bibinfo{author}{Brockhoff, P.~B.},  and
  \bibinfo{author}{Christensen, R. H.~B.} (\textbf{\bibinfo{year}{2017}}).
  \enquote{\bibinfo{title}{{lmerTest} package: Tests in linear mixed effects
  models}} \bibinfo{journal}{J. Statist. Softw.} \textbf{82}(13),
  \bibinfo{pages}{1--26}.

\bibitem[{Lammert \emph{et~al.}(2013{\natexlab{a}})Lammert, Proctor, and
  Narayanan}]{lammert2013b}
\bibinfo{author}{Lammert, A.}, \bibinfo{author}{Proctor, M.},  and
  \bibinfo{author}{Narayanan, S.}
  (\textbf{\bibinfo{year}{2013}}{\natexlab{a}}).
  \enquote{\bibinfo{title}{Interspeaker variability in hard palate morphology
  and vowel production}} \bibinfo{journal}{J. Speech Lang. Hear. Res.}
  \textbf{56}(6), \bibinfo{pages}{1924--1933}.

\bibitem[{Lammert \emph{et~al.}(2013{\natexlab{b}})Lammert, Proctor, and
  Narayanan}]{lammert2013a}
\bibinfo{author}{Lammert, A.}, \bibinfo{author}{Proctor, M.},  and
  \bibinfo{author}{Narayanan, S.}
  (\textbf{\bibinfo{year}{2013}}{\natexlab{b}}).
  \enquote{\bibinfo{title}{Morphological variation in the adult hard palate and
  posterior pharyngeal wall}} \bibinfo{journal}{J. Speech Lang. Hear. Res.}
  \textbf{56}(2), \bibinfo{pages}{521--530}.

\bibitem[{Laver(1980)}]{laver1980}
\bibinfo{author}{Laver, J.} (\textbf{\bibinfo{year}{1980}}).
  \emph{\bibinfo{title}{The Phonetic Description of Voice Quality}}
  (\bibinfo{publisher}{Cambridge University Press},
  \bibinfo{address}{Cambridge, UK}).

\bibitem[{Lisker(1985)}]{lisker1985}
\bibinfo{author}{Lisker, L.} (\textbf{\bibinfo{year}{1985}}).
  \enquote{\bibinfo{title}{The pursuit of invariance in speech signals}}
  \bibinfo{journal}{J. Acoust. Soc. Am.} \textbf{77}(3),
  \bibinfo{pages}{1199--1202}.

\bibitem[{Lo \emph{et~al.}(2025)Lo, Strycharczuk, and Kirkham}]{lo2025}
\bibinfo{author}{Lo, J. J.~H.}, \bibinfo{author}{Strycharczuk, P.},  and
  \bibinfo{author}{Kirkham, S.} (\textbf{\bibinfo{year}{2025}}).
  \enquote{\bibinfo{title}{Articulatory strategy in vowel production as a basis
  for speaker discrimination}} in \emph{\bibinfo{booktitle}{Proc. Interspeech
  2025}}, pp. \bibinfo{pages}{3504--3508}.

\bibitem[{Mathis \emph{et~al.}(2018)Mathis, Mamidanna, Cury, Abe, Murthy,
  Mathis, and Bethge}]{mathis2018}
\bibinfo{author}{Mathis, A.}, \bibinfo{author}{Mamidanna, P.},
  \bibinfo{author}{Cury, K.~M.}, \bibinfo{author}{Abe, T.},
  \bibinfo{author}{Murthy, V.~N.}, \bibinfo{author}{Mathis, M.~W.},  and
  \bibinfo{author}{Bethge, M.} (\textbf{\bibinfo{year}{2018}}).
  \enquote{\bibinfo{title}{{DeepLabCut}: markerless pose estimation of
  user-defined body parts with deep learning}} \bibinfo{journal}{Nature
  Neurosci.} \textbf{21}(9), \bibinfo{pages}{1281--1289}.

\bibitem[{McAuliffe \emph{et~al.}(2017)McAuliffe, Socolof, Mihuc, Wagner, and
  Sonderegger}]{mcauliffe17}
\bibinfo{author}{McAuliffe, M.}, \bibinfo{author}{Socolof, M.},
  \bibinfo{author}{Mihuc, S.}, \bibinfo{author}{Wagner, M.},  and
  \bibinfo{author}{Sonderegger, M.} (\textbf{\bibinfo{year}{2017}}).
  \enquote{\bibinfo{title}{Montreal {F}orced {A}ligner: {T}rainable text-speech
  alignment using {K}aldi}} in \emph{\bibinfo{booktitle}{Proc. Interspeech
  2017}}, pp. \bibinfo{pages}{498--502}.

\bibitem[{McDougall(2004)}]{mcdougall2004}
\bibinfo{author}{McDougall, K.} (\textbf{\bibinfo{year}{2004}}).
  \enquote{\bibinfo{title}{Speaker-specific formant dynamics: An experiment on
  {A}ustralian {E}nglish /\textipa{AI}/}} \bibinfo{journal}{Int. J. Speech
  Lang. Law} \textbf{11}(1), \bibinfo{pages}{103--130}.

\bibitem[{McDougall(2006)}]{mcdougall2006}
\bibinfo{author}{McDougall, K.} (\textbf{\bibinfo{year}{2006}}).
  \enquote{\bibinfo{title}{Dynamic features of speech and the characterization
  of speakers: Toward a new approach using formant frequencies}}
  \bibinfo{journal}{Int. J. Speech Lang. Law} \textbf{13}(1),
  \bibinfo{pages}{89--126}.

\bibitem[{McDougall and Nolan(2007)}]{mcdougall2007}
\bibinfo{author}{McDougall, K.},  and \bibinfo{author}{Nolan, F.}
  (\textbf{\bibinfo{year}{2007}}). \enquote{\bibinfo{title}{Discrimination of
  speakers using the formant dynamics of /\textipa{u:}/ in {B}ritish
  {E}nglish}} in \emph{\bibinfo{booktitle}{Proc. 16th Inter. Congr. Phonetic
  Sci.}}, edited by \bibinfo{editor}{J.~Trouvain} and \bibinfo{editor}{W.~J.
  Barry}, pp. \bibinfo{pages}{1825--1828}.

\bibitem[{McFarland and Baum(1995)}]{mcfarland1995}
\bibinfo{author}{McFarland, D.~H.},  and \bibinfo{author}{Baum, S.~R.}
  (\textbf{\bibinfo{year}{1995}}). \enquote{\bibinfo{title}{Incomplete
  compensation to articulatory perturbation}} \bibinfo{journal}{J. Acoust. Soc.
  Am.} \textbf{97}(3), \bibinfo{pages}{1865--1873}.

\bibitem[{Morrison(2009)}]{morrison2009}
\bibinfo{author}{Morrison, G.~S.} (\textbf{\bibinfo{year}{2009}}).
  \enquote{\bibinfo{title}{Likelihood-ratio forensic voice comparison using
  parametric representations of the formant trajectories of diphthongs}}
  \bibinfo{journal}{J. Acoust. Soc. Am.} \textbf{125}(4),
  \bibinfo{pages}{2387--2397}.

\bibitem[{Noiray \emph{et~al.}(2014)Noiray, Iskarous, and Whalen}]{noiray2014}
\bibinfo{author}{Noiray, A.}, \bibinfo{author}{Iskarous, K.},  and
  \bibinfo{author}{Whalen, D.} (\textbf{\bibinfo{year}{2014}}).
  \enquote{\bibinfo{title}{Variability in {E}nglish vowels is comparable in
  articulation and acoustics}} \bibinfo{journal}{Lab. Phonol.} \textbf{5}(2),
  \bibinfo{pages}{271--288}.

\bibitem[{Nolan(1983)}]{nolan1983}
\bibinfo{author}{Nolan, F.} (\textbf{\bibinfo{year}{1983}}).
  \emph{\bibinfo{title}{The Phonetic Bases of Speaker Recognition}}
  (\bibinfo{publisher}{Cambridge University Press},
  \bibinfo{address}{Cambridge, UK}).

\bibitem[{Nolan and Grigoras(2005)}]{nolan2005}
\bibinfo{author}{Nolan, F.},  and \bibinfo{author}{Grigoras, C.}
  (\textbf{\bibinfo{year}{2005}}). \enquote{\bibinfo{title}{A case for formant
  analysis in forensic speaker identification}} \bibinfo{journal}{Int. J.
  Speech Lang. Law} \textbf{12}(2), \bibinfo{pages}{143--173}.

\bibitem[{Redford and Baese-Berk(2023)}]{redford2023}
\bibinfo{author}{Redford, M.},  and \bibinfo{author}{Baese-Berk, M.}
  (\textbf{\bibinfo{year}{2023}}). \enquote{\bibinfo{title}{Acoustic theories
  of speech perception}} in \emph{\bibinfo{booktitle}{Oxford Research
  Encyclopedia of Linguistics}}.

\bibitem[{Rhodes(2012)}]{rhodes2012}
\bibinfo{author}{Rhodes, R.~W.} (\textbf{\bibinfo{year}{2012}}).
  \enquote{\bibinfo{title}{Assessing the strength of non-contemporaneous
  forensic speech evidence}} Ph.D. thesis, \bibinfo{school}{University of
  York}.

\bibitem[{Rose \emph{et~al.}(2006)Rose, Warren, and Watson}]{rose2006}
\bibinfo{author}{Rose, P.}, \bibinfo{author}{Warren, P.},  and
  \bibinfo{author}{Watson, C.} (\textbf{\bibinfo{year}{2006}}).
  \enquote{\bibinfo{title}{The intrinsic forensic discriminatory power of
  diphthongs}} in \emph{\bibinfo{booktitle}{Proc. 11th Aust. Int. Conf. Speech
  Sci. Technol.}}, pp. \bibinfo{pages}{64--69}.

\bibitem[{Saltzman and Munhall(1989)}]{saltzman1989}
\bibinfo{author}{Saltzman, E.~L.},  and \bibinfo{author}{Munhall, K.~G.}
  (\textbf{\bibinfo{year}{1989}}). \enquote{\bibinfo{title}{A dynamical
  approach to gestural patterning in speech production}}
  \bibinfo{journal}{Ecol. Psychol.} \textbf{1}(4), \bibinfo{pages}{333--382}.

\bibitem[{Serrurier \emph{et~al.}(2019)Serrurier, Badin, Lamalle, and
  Neuschaefer-Rube}]{serrurier2019}
\bibinfo{author}{Serrurier, A.}, \bibinfo{author}{Badin, P.},
  \bibinfo{author}{Lamalle, L.},  and \bibinfo{author}{Neuschaefer-Rube, C.}
  (\textbf{\bibinfo{year}{2019}}). \enquote{\bibinfo{title}{Characterization of
  inter-speaker articulatory variability: A two-level multi-speaker modelling
  approach based on {MRI} data}} \bibinfo{journal}{J. Acoust. Soc. Am.}
  \textbf{145}(4), \bibinfo{pages}{2149--2170}.

\bibitem[{Serrurier and Neuschaefer-Rube(2023)}]{serrurier2023}
\bibinfo{author}{Serrurier, A.},  and \bibinfo{author}{Neuschaefer-Rube, C.}
  (\textbf{\bibinfo{year}{2023}}). \enquote{\bibinfo{title}{Morphological and
  acoustic modeling of the vocal tract}} \bibinfo{journal}{J. Acoust. Soc. Am.}
  \textbf{153}(3), \bibinfo{pages}{1867--1886}.

\bibitem[{Serrurier and Neuschaefer-Rube(2024)}]{serrurier2024}
\bibinfo{author}{Serrurier, A.},  and \bibinfo{author}{Neuschaefer-Rube, C.}
  (\textbf{\bibinfo{year}{2024}}). \enquote{\bibinfo{title}{Formant-based
  articulatory strategies: {C}haracterisation and inter-speaker variability
  analysis}} \bibinfo{journal}{J. Phonetics} \textbf{107},
  \bibinfo{pages}{101374}.

\bibitem[{S{\'o}skuthy(2021)}]{soskuthy2021}
\bibinfo{author}{S{\'o}skuthy, M.} (\textbf{\bibinfo{year}{2021}}).
  \enquote{\bibinfo{title}{Evaluating generalised additive mixed modelling
  strategies for dynamic speech analysis}} \bibinfo{journal}{J. Phonetics}
  \textbf{84}, \bibinfo{pages}{101017}.

\bibitem[{Spreafico \emph{et~al.}(2018)Spreafico, Pucher, and
  Matosova}]{spreafico2018}
\bibinfo{author}{Spreafico, L.}, \bibinfo{author}{Pucher, M.},  and
  \bibinfo{author}{Matosova, A.} (\textbf{\bibinfo{year}{2018}}).
  \enquote{\bibinfo{title}{Ultra{F}it: {A} speaker-friendly headset for
  ultrasound recordings in speech science}} in \emph{\bibinfo{booktitle}{Proc.
  Interspeech 2018}}, \bibinfo{organization}{International Speech Communication
  Association}, pp. \bibinfo{pages}{1517--1520}.

\bibitem[{Stevens(1989)}]{stevens1989}
\bibinfo{author}{Stevens, K.~N.} (\textbf{\bibinfo{year}{1989}}).
  \enquote{\bibinfo{title}{On the quantal nature of speech}}
  \bibinfo{journal}{J. Phonetics} \textbf{17}(1--2), \bibinfo{pages}{3--45}.

\bibitem[{Strycharczuk and Kirkham(2025)}]{strycharczuk_gender}
\bibinfo{author}{Strycharczuk, P.},  and \bibinfo{author}{Kirkham, S.}
  (\textbf{\bibinfo{year}{2025}}). \enquote{\bibinfo{title}{Articulatory
  strategies in male and female vowel production}} \bibinfo{journal}{J. Speech
  Lang. Hear. Res.} \textbf{68}(12), \bibinfo{pages}{5629--5649}.

\bibitem[{Strycharczuk \emph{et~al.}(2024)Strycharczuk, Kirkham, Gorman, and
  Nagamine}]{strycharczuk2024_towards}
\bibinfo{author}{Strycharczuk, P.}, \bibinfo{author}{Kirkham, S.},
  \bibinfo{author}{Gorman, E.},  and \bibinfo{author}{Nagamine, T.}
  (\textbf{\bibinfo{year}{2024}}). \enquote{\bibinfo{title}{Towards a dynamical
  model of {E}nglish vowels. {E}vidence from diphthongisation}}
  \bibinfo{journal}{J. Phonetics} \textbf{107}, \bibinfo{pages}{101349}.

\bibitem[{Strycharczuk \emph{et~al.}(2025)Strycharczuk, Kirkham, Gorman, and
  Nagamine}]{strycharczuk2025_dimensionality}
\bibinfo{author}{Strycharczuk, P.}, \bibinfo{author}{Kirkham, S.},
  \bibinfo{author}{Gorman, E.},  and \bibinfo{author}{Nagamine, T.}
  (\textbf{\bibinfo{year}{2025}}). \enquote{\bibinfo{title}{Dimensionality
  reduction in lingual articulation of vowels: {E}vidence from lax vowels in
  {N}orthern {A}nglo-{E}nglish}} \bibinfo{journal}{Lang. Speech}
  \textbf{68}(3), \bibinfo{pages}{689--721}.

\bibitem[{Strycharczuk \emph{et~al.}(2020)Strycharczuk,
  L{\'o}pez-Ib{\'a}{\~n}ez, Brown, and Leemann}]{strycharczuk2020}
\bibinfo{author}{Strycharczuk, P.}, \bibinfo{author}{L{\'o}pez-Ib{\'a}{\~n}ez,
  M.}, \bibinfo{author}{Brown, G.},  and \bibinfo{author}{Leemann, A.}
  (\textbf{\bibinfo{year}{2020}}). \enquote{\bibinfo{title}{General {N}orthern
  {E}nglish. {E}xploring regional variation in the {N}orth of {E}ngland with
  machine learning}} \bibinfo{journal}{Frontiers Artif. Intell.} \textbf{3},
  \bibinfo{pages}{48}.

\bibitem[{Watson and Harrington(1999)}]{watson1999}
\bibinfo{author}{Watson, C.~I.},  and \bibinfo{author}{Harrington, J.}
  (\textbf{\bibinfo{year}{1999}}). \enquote{\bibinfo{title}{Acoustic evidence
  for dynamic formant trajectories in {A}ustralian {E}nglish vowels}}
  \bibinfo{journal}{J. Acoust. Soc. Am.} \textbf{106}(1),
  \bibinfo{pages}{458--468}.

\bibitem[{Watt(2002)}]{watt2002}
\bibinfo{author}{Watt, D.} (\textbf{\bibinfo{year}{2002}}).
  \enquote{\bibinfo{title}{‘{I} don’t speak with a {G}eordie accent, {I}
  speak, like, the {N}orthern accent’: {C}ontact-induced levelling in the
  {T}yneside vowel system}} \bibinfo{journal}{J. Sociolinguistics}
  \textbf{6}(1), \bibinfo{pages}{44--63}.

\bibitem[{Weirich(2012)}]{weirich2012}
\bibinfo{author}{Weirich, M.} (\textbf{\bibinfo{year}{2012}}).
  \enquote{\bibinfo{title}{The influence of {N}ature and {N}urture on
  speaker-specific parameters in twins speech}} Ph.D. thesis,
  \bibinfo{school}{Humboldt-Universit{\"a}t zu Berlin}.

\bibitem[{Weirich \emph{et~al.}(2016)Weirich, Fuchs, Simpson, Winkler, and
  Perrier}]{weirich2016}
\bibinfo{author}{Weirich, M.}, \bibinfo{author}{Fuchs, S.},
  \bibinfo{author}{Simpson, A.}, \bibinfo{author}{Winkler, R.},  and
  \bibinfo{author}{Perrier, P.} (\textbf{\bibinfo{year}{2016}}).
  \enquote{\bibinfo{title}{Mumbling: Macho or morphology?}}
  \bibinfo{journal}{J. Speech Lang. Hear. Res.} \textbf{59}(6),
  \bibinfo{pages}{S1587--S1595}.

\bibitem[{Weirich and Simpson(2018)}]{weirich2018individual}
\bibinfo{author}{Weirich, M.},  and \bibinfo{author}{Simpson, A.~P.}
  (\textbf{\bibinfo{year}{2018}}). \enquote{\bibinfo{title}{Individual
  differences in acoustic and articulatory undershoot in a {G}erman
  diphthong--variation between male and female speakers}} \bibinfo{journal}{J.
  Phonetics} \textbf{71}, \bibinfo{pages}{35--50}.

\bibitem[{Wells(1982)}]{wells1982.1}
\bibinfo{author}{Wells, J.} (\textbf{\bibinfo{year}{1982}}).
  \emph{\bibinfo{title}{Accents of {E}nglish 1: An introduction}},
  \bibinfo{volume}{\textbf{2}} (\bibinfo{publisher}{Cambridge University
  Press}, \bibinfo{address}{Camrbidge, UK}).

\bibitem[{Wood(2017)}]{wood2017}
\bibinfo{author}{Wood, S.} (\textbf{\bibinfo{year}{2017}}).
  \emph{\bibinfo{title}{Generalized Additive Models: An Introduction with
  {R}}}, \bibinfo{edition}{2nd} ed. (\bibinfo{publisher}{Chapman and
  Hall/CRC}).

\bibitem[{Wrench and Balch-Tomes(2022)}]{wrench2022}
\bibinfo{author}{Wrench, A.},  and \bibinfo{author}{Balch-Tomes, J.}
  (\textbf{\bibinfo{year}{2022}}). \enquote{\bibinfo{title}{Beyond the edge:
  {M}arkerless pose estimation of speech articulators from ultrasound and
  camera images using {D}eep{L}ab{C}ut}} \bibinfo{journal}{Sensors}
  \textbf{22}(3), \bibinfo{pages}{1133}.

\end{thebibliography}

\end{document}